\documentclass{egpubl}
\usepackage{gch2022}
 
 \WsShortPaper      % uncomment for Short Paper contribution
\usepackage[T1]{fontenc}
\usepackage{dfadobe}  

\usepackage{cite}  % comment out for biblatex with backend=biber
\BibtexOrBiblatex
\electronicVersion
\PrintedOrElectronic
\ifpdf \usepackage[pdftex]{graphicx} \pdfcompresslevel=9
\else \usepackage[dvips]{graphicx} \fi

\usepackage{egweblnk} 
\title[Reconstructing Stucco Statues from historic Sketches]%
      {A Concept for Reconstructing Stucco Statues from historic Sketches using synthetic Data only}
\author[Thomas Pöllabauer \& Julius Kühn]
{\parbox{\textwidth}{\centering T. Pöllabauer$^{1,2}$\orcid{0000-0003-0075-1181}
        and J. Kühn$^{1}$\orcid{0000-0003-2458-0030} 
        }
        \\
{\parbox{\textwidth}{\centering $^1$Fraunhofer Institute for Computer Graphics Research IGD \\
                                $^2$Interactive Graphics Research Group, TU Darmstadt
      }
}
}

\begin{document}

\teaser{
 \includegraphics[width=0.8\linewidth]{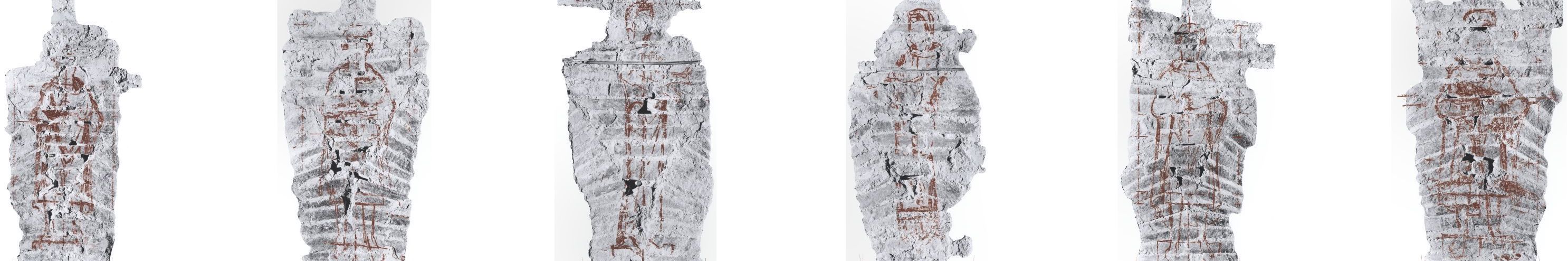}
 \centering
  \caption{Historic sketches used for stucco work as found in the Princely Abbey of Corvey. Illustrations as found in \cite{teaser}.}
\label{fig:teaser}
}

\maketitle
%-------------------------------------------------------------------------
\begin{abstract}
  %The problem: 
  In medieval times, stuccoworkers used a red color, called sinopia, to first create a sketch of the to-be-made statue on the wall. %Why the problem is interesting: 
  Today, many of these statues are destroyed, but using the original drawings, deriving from the red color also called sinopia, we can reconstruct how the final statue might have looked. %What our solution achieves: 
  We propose a fully-automated approach to reconstruct a point cloud and show preliminary results by generating a color-image, a depth-map, as well as surface normals requiring only a single sketch, and without requiring a collection of other, similar samples. %What is the consequence of our solution:
  Our proposed solution allows real-time reconstruction on-site, for instance, within an exhibition, or to generate a useful starting point for an expert, trying to manually reconstruct the statue, all while using only synthetic data for training. 
%-------------------------------------------------------------------------
%  ACM CCS 1998
%  (see https://www.acm.org/publications/computing-classification-system/1998)
% \begin{classification} % according to https://www.acm.org/publications/computing-classification-system/1998
% \CCScat{Computer Graphics}{I.3.3}{Picture/Image Generation}{Line and curve generation}
% \end{classification}
%-------------------------------------------------------------------------
%  ACM CCS 2012
%  (see https://www.acm.org/publications/class-2012)
%The tool at \url{http://dl.acm.org/ccs.cfm} can be used to generate
% CCS codes.
%Example:


\end{abstract}  
%-------------------------------------------------------------------------

\section{Introduction}
In 1992, during an inspection of the stonework in the westwork of the Princely Abbey of Corvey, oxide red brushstrokes were unexpectedly found \cite{corvey}. It turned out, these brushstrokes belonged to one of six wall drawings, called sinopia, depicting four men, and two women. However, it soon became clear that these drawings were not meant to be the preliminary stage of a painting, but of stucco statues, as proved by wooden stakes driven into the wall and small residues of material containing gypsum, around these stakes. In addition, stucco fragments found in 1961 proved to be a match for the newly found sinopia. Given the wall drawings and the few remaining fragments, interest arose in reconstructing the destroyed statues. \\
While with only six drawings, one can manually do the reconstruction, an automatable approach could be the basis for a more general solution, also applicable to other historic drawings, found at other locations. Ideally, a solution would not require a domain expert at every part of the process and works with only little to no real data for parameterization, since real samples tend to be scarce and vary drastically in their conservation state and details. \\
We propose a data driven approach to (semi-)automatically produce a first estimate of a destroyed statue, based on a degraded-by-time line drawing. We present the outline of our proposed end-to-end pipeline, as well as first results, demonstrating the feasibility of reconstructing without real data. Our contributions are: 
\begin{itemize}
  \item An end-to-end pipeline for geometry reconstruction of historic statues and figurines based on sketches from the period.
  \item As a test for feasibility, given a simple line drawing of a historic sketch, we generate a depth estimation, full-color reconstruction, as well as surface normals.
  \item We train on synthetic data, i.e. we do not need any other sketches, neither from the time period, nor other, thereby solving the prevailing problem in cultural heritage, of not having enough data for machine learning approaches.
  \item By use of abstraction, our approach is applicable not only to sinopia.
  \item Aside of initial sketch restoration, we do not incorporate domain knowledge, making the solution more general.
\end{itemize}  

%Overview of our proposed approach. We solve the problem in 3 stages: First, we collect unrelated statues depicting humans, data that is freely available in large quantities. We use this data to generate images, which we reduce to mere line drawings, making it less distinguishable from our real samples. Second, we train a highly expressive deep neural network to reconstruct high quality information from the reduced representation. Third, we apply our image translation to our handful of real-world samples before estimating the 3d shape. To make the pipeline end-to-end learn-able, the mapping function between 2d-3d information should ideally be differentiable. We suggest the use of a differentiable renderer such as \URL{https://pytorch3d.org}.}

\begin{figure}
  \centering
  \mbox{}
  \includegraphics[width=0.48\textwidth]{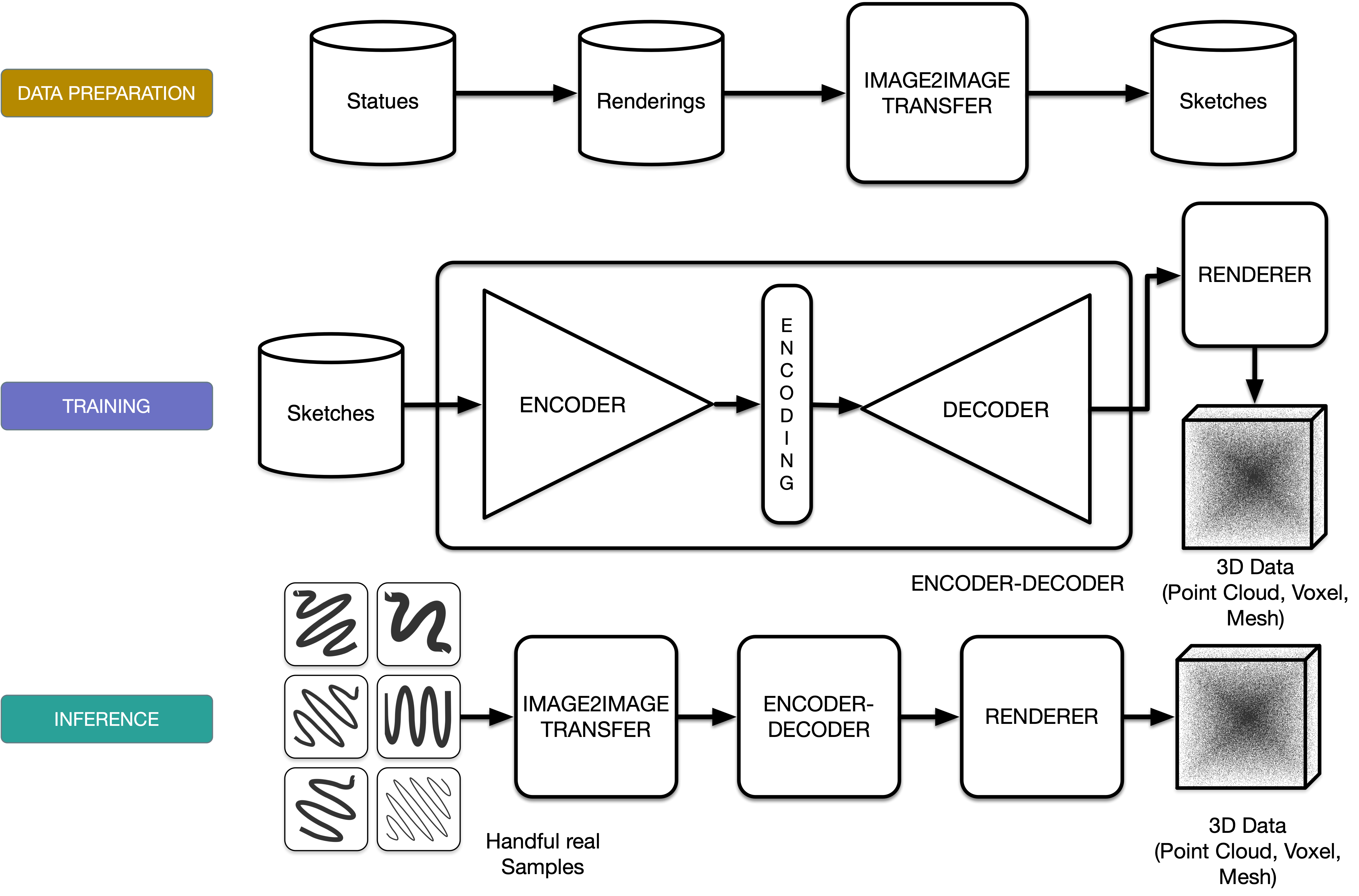}
  \caption{\label{fig:Overview} Overview of our approach. We solve the problem in 3 stages: First, we collect unrelated statues depicting humans. We use this data to generate images, which we reduce to mere line drawings, making it less distinguishable from our real samples. Second, we train a highly expressive deep neural network to reconstruct high quality information from the reduced representation. Third, we apply our image translation to our handful of real-world samples before estimating the 3d shape. To make the pipeline end-to-end learn-able, the mapping function between 2d-3d information should ideally be differentiable. }
\end{figure}

\section{Related Work}
Geometry reconstruction based on images is a highly relevant challenge when trying to preserve cultural artifacts and locales. One application is scanning using sophisticated sensors, such as laser or RGB(+D) sensors together with photogrammetry. \cite{reconstructionSurvey} provides a survey of these and similar methods. A major drawback of these approaches is the requirement of existing geometry. While we have some remaining fragments, these represent so little of the original statue's volume that we cannot reconstruct the full statue relying only on these fragments, using a technique as described in \cite{3dReconFromDefectiveObjects} for example. Instead, we have to rely on a single depiction, the sketch on the wall, to reconstruct the appearance. \\
Reconstructing 3d information from images is a long withstanding problem in the computer vision community. It is an inherently ill-posed problem since the imaging process, by design, forfeits knowledge about the depicted scene. One way of dealing with this loss of information is the use of multiple images to reconstruct the 3d information. The availability of additional view points reintroduces much of the lost information about the scene. However, the need for multiple images of a single scene greatly reduces the applicability in real world scenarios by requiring either a multi-camera setup or a static scene, photographed from various view points. This constraint led to interest in solving the reconstruction problem via a single image only. Solving the problem with only one view requires the introduction of additional information, for instance knowledge about objects depicted in the scene. \\
Another possible solution, which proved very successful recently, is the use of machine learning and especially deep neural networks. These algorithms can be trained with a wide variety of scenes with available ground truth data and thereby acquire some general scene understanding or, in other words, a useful prior on the basic construction of scenes and their projection on a 2-dimensional image plane \cite{LearningPCDGen}\cite{P2Vpp}. These approaches often lead to impressive results given the ill posed problem, but usually require real world photographs or realistic imagery, such as renderings. Also, in the cultural heritage domain, current deep learning approaches face the problem of too little data, as noted in \cite{ML4CHsurvey}, surveying the use of machine learning in cultural heritage. \\
There is a much smaller body of work dealing with geometry estimation based on hand-drawn sketches of various degrees of complexity \cite{3DRecFromSingleSketch}. These approaches might use segmentation \cite{3DRecFromSingleSketch} and/or estimate depth and normals before fusing the results to different kind of 3d representations \cite{multi-view-sketch-recon}.\\
Some works, especially those relying on differentiable rendering, do not require anything, but geometric understanding of the imaging process \cite{largesteps}\cite{SDFDiff}. In our data limited scenario, however, we favor approaches that allow to learn a prior on other, similar, available data. A often-utilized approach is the use of encoder-decoder architectures.
Encoder-decoder architectures break an input image (usually using a convolutional neural network) down into a, compared to the original image, small vector representation. This vector space is called the latent space. A second network, the decoder samples vectors from this latent space and has to reconstruct the encoded image. This allows to learn a powerful representation on available training data to be used in cases with only a few real samples, an ideal fit to our challenge.

\section{Approach}

Our proposed solution takes a sketch of a sinopia. As shown in row 1 of Fig. \ref{fig:teaser}, the drawing is almost unobservable, which is why we require help of a domain expert to extract the contours and fill in missing details. This is the only time we require domain knowledge. Next, we use image-to-image-translation to project this sketch in our intermediary domain of line drawings. Finally, we pass the converted image to an encoder-decoder network, pre-trained on a dataset of statues with geometry information available. Our approach is depicted in Fig. \ref{fig:Overview}.
\begin{table*}[h]
\centering
\begin{tabular}{ |p{0.2cm}|p{1.2cm}||p{1.2cm}|p{1.2cm}|p{1.2cm}|p{1.2cm}|p{0.2cm}|p{1.2cm}||p{1.2cm}|p{1.2cm}|p{1.2cm}|p{1.2cm}| }
	\hline
	\multicolumn{12}{|c|}{Results on Test Split} \\
	\hline
	No & Sketch & RGB & Depth & Normals & Mask & No & Sketch & RGB & Depth &  Normals &  Mask \\
	\hline
	1 & \raisebox{-.5\height}{\includegraphics[height=14mm,width=14mm,keepaspectratio]{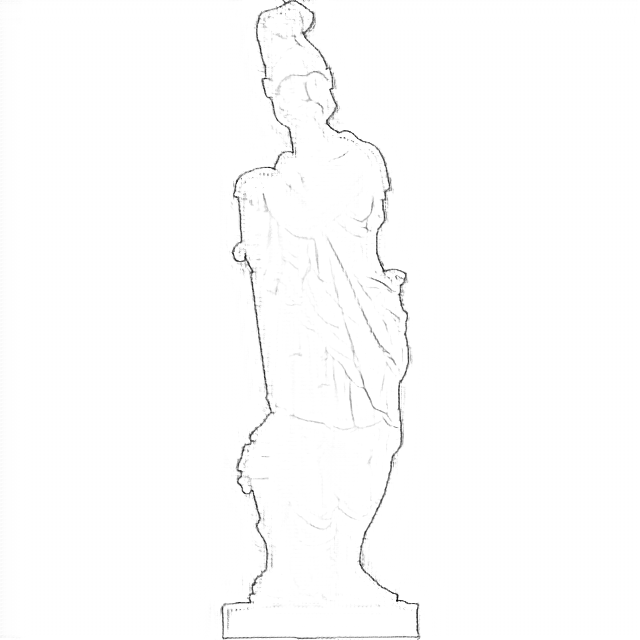}} &  \raisebox{-.5\height}{\includegraphics[height=14mm,width=14mm,keepaspectratio]{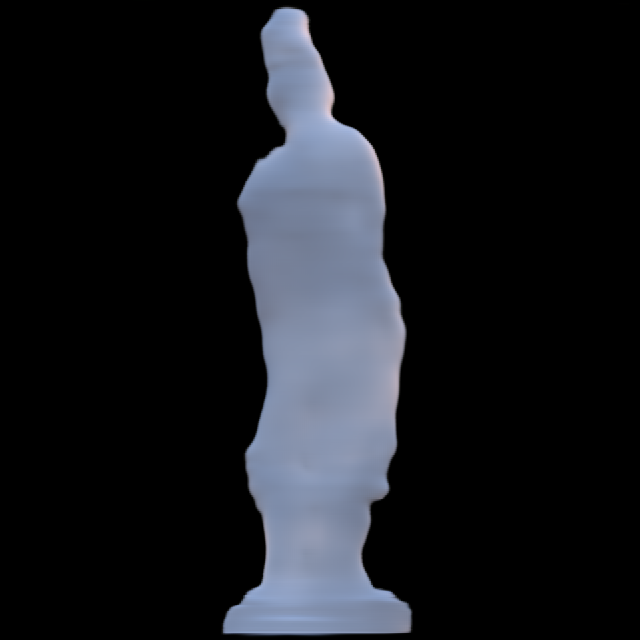}} & \raisebox{-.5\height}{\includegraphics[height=14mm,width=14mm,keepaspectratio]{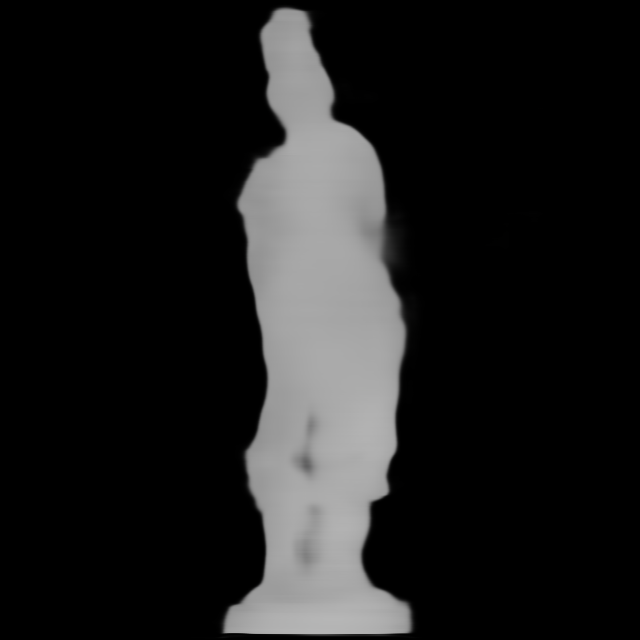}} & \raisebox{-.5\height}{\includegraphics[height=14mm,width=14mm,keepaspectratio]{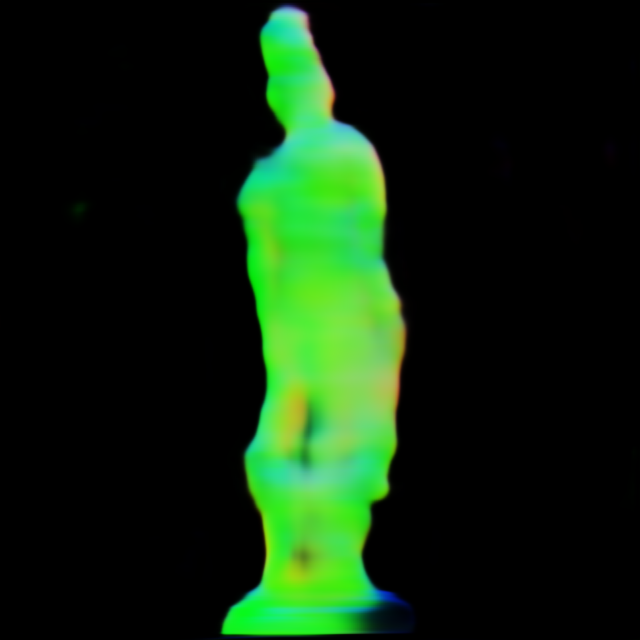}} & \raisebox{-.5\height}{\includegraphics[height=14mm,width=14mm,keepaspectratio]{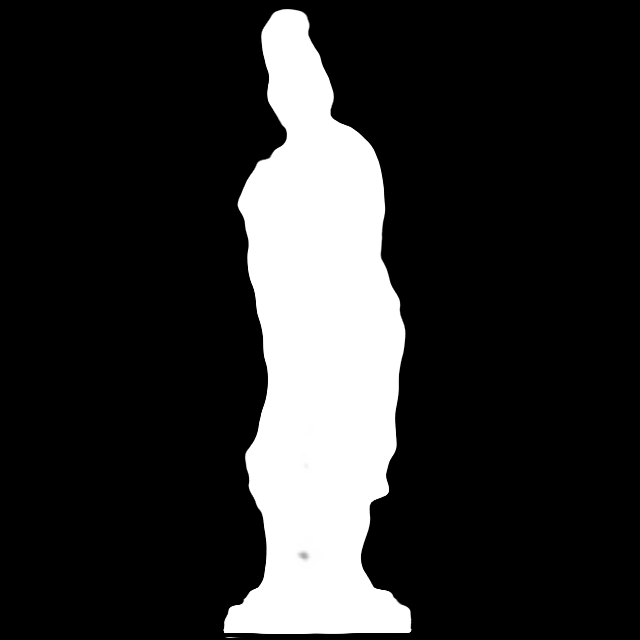}} &
	2 & \raisebox{-.5\height}{\includegraphics[height=14mm,width=14mm,keepaspectratio]{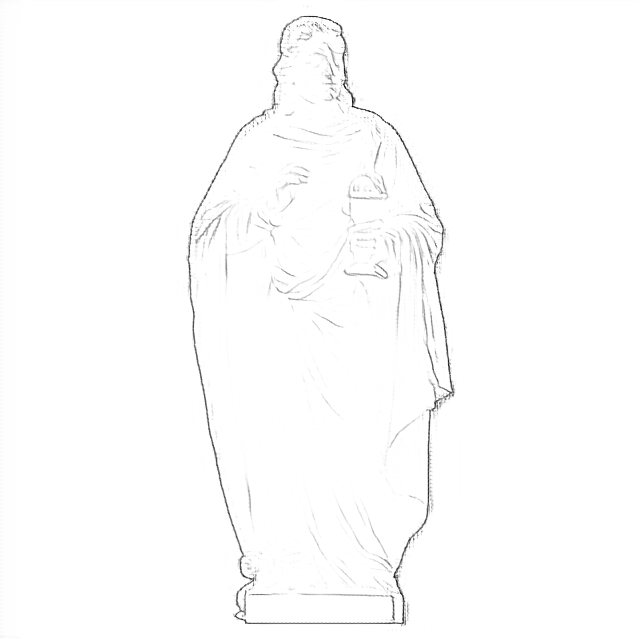}} &  \raisebox{-.5\height}{\includegraphics[height=14mm,width=14mm,keepaspectratio]{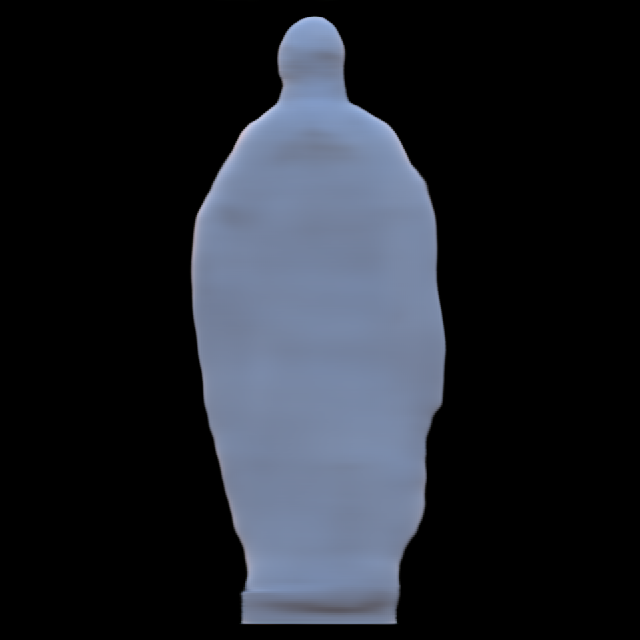}} & \raisebox{-.5\height}{\includegraphics[height=14mm,width=14mm,keepaspectratio]{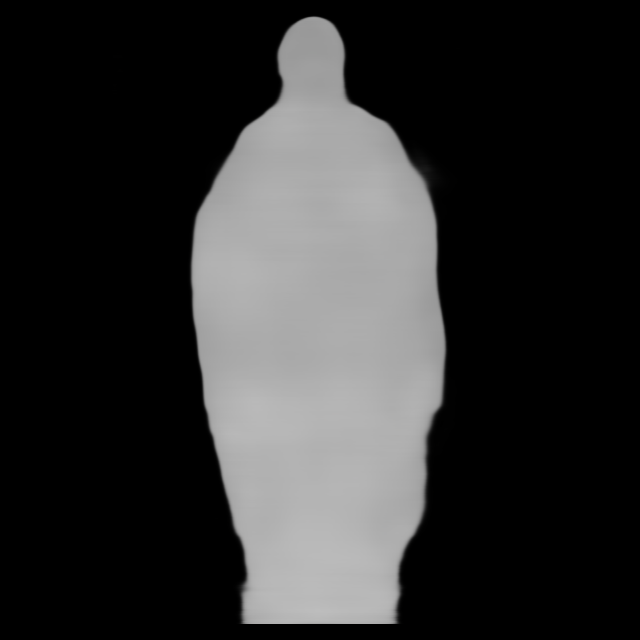}} & \raisebox{-.5\height}{\includegraphics[height=14mm,width=14mm,keepaspectratio]{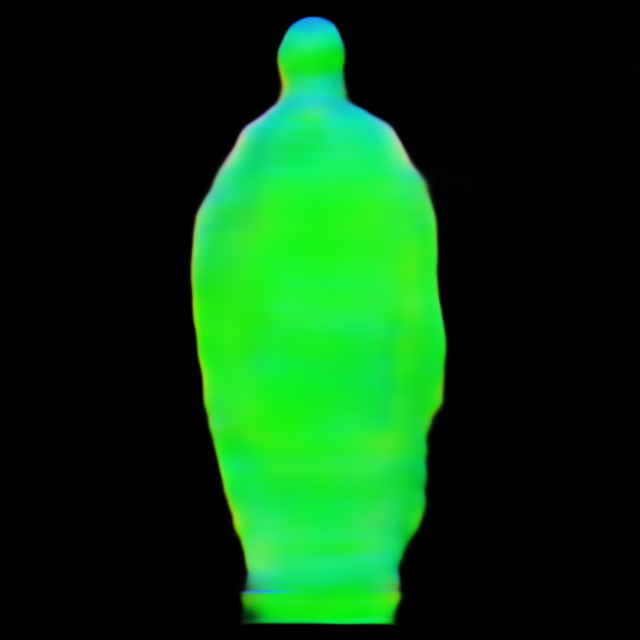}} & \raisebox{-.5\height}{\includegraphics[height=14mm,width=14mm,keepaspectratio]{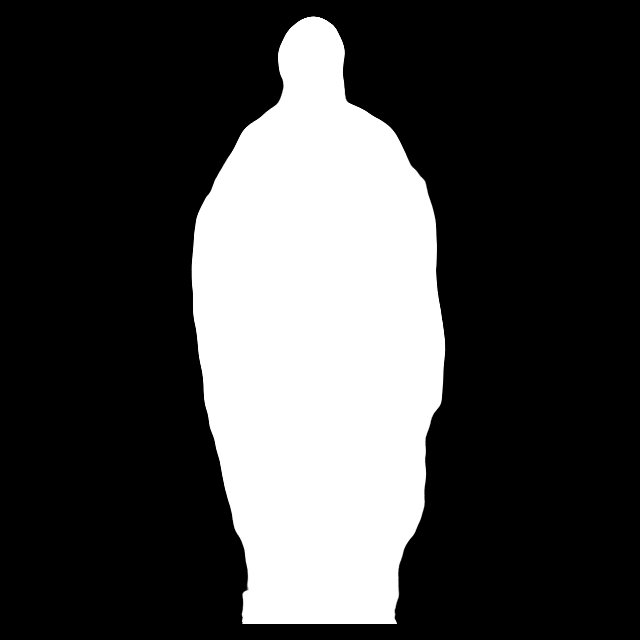}}   \\
	3 & \raisebox{-.5\height}{\includegraphics[height=14mm,width=14mm,keepaspectratio]{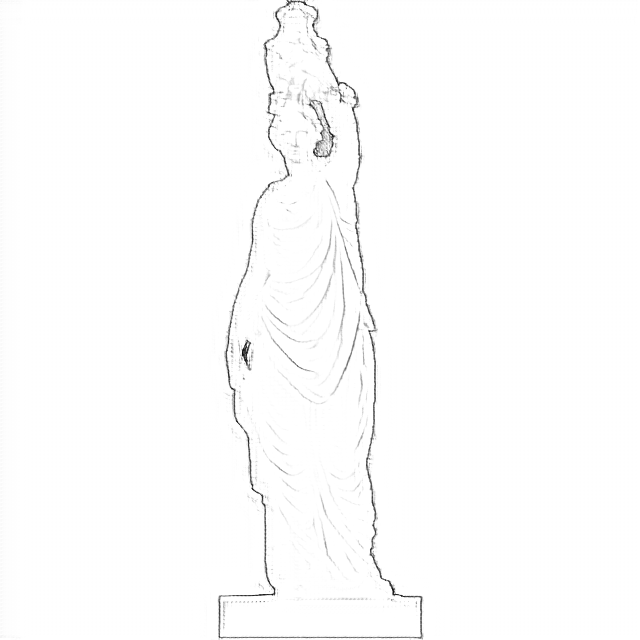}} &  \raisebox{-.5\height}{\includegraphics[height=14mm,width=14mm,keepaspectratio]{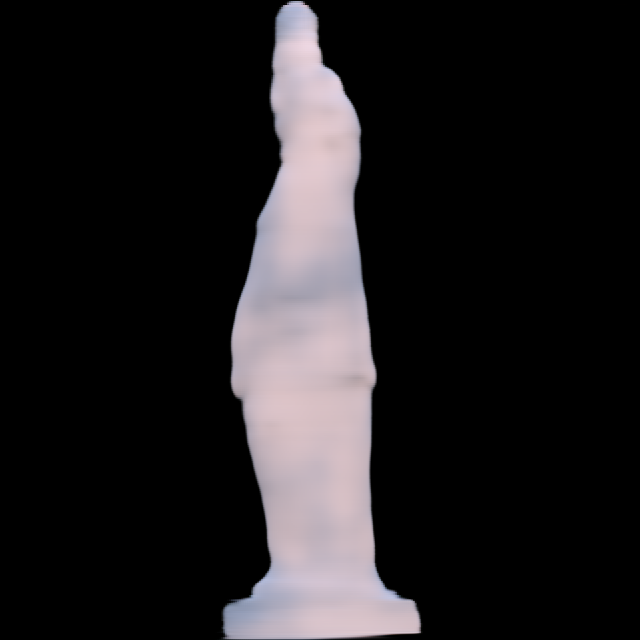}} & \raisebox{-.5\height}{\includegraphics[height=14mm,width=14mm,keepaspectratio]{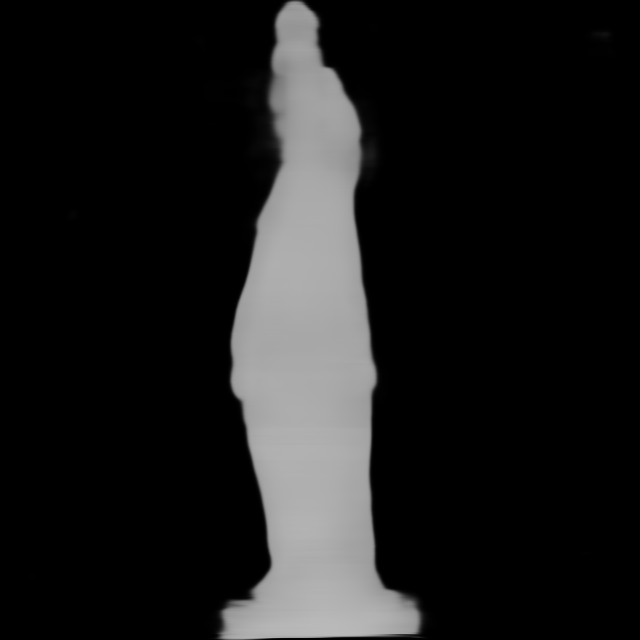}} & \raisebox{-.5\height}{\includegraphics[height=14mm,width=14mm,keepaspectratio]{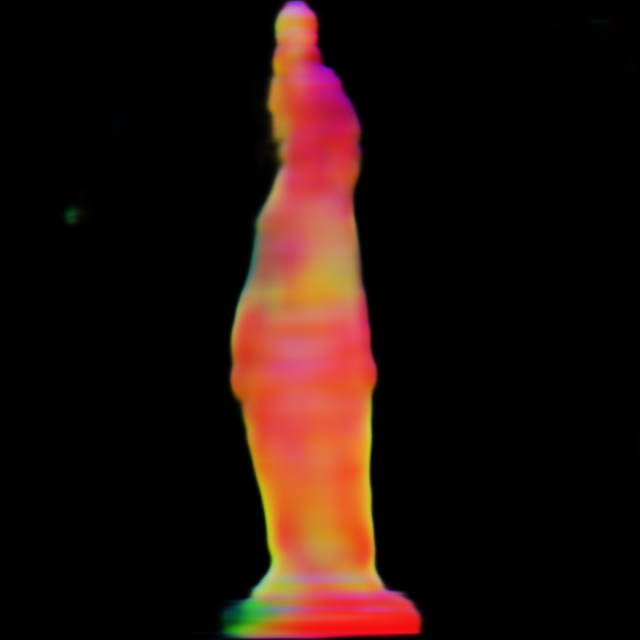}} & \raisebox{-.5\height}{\includegraphics[height=14mm,width=14mm,keepaspectratio]{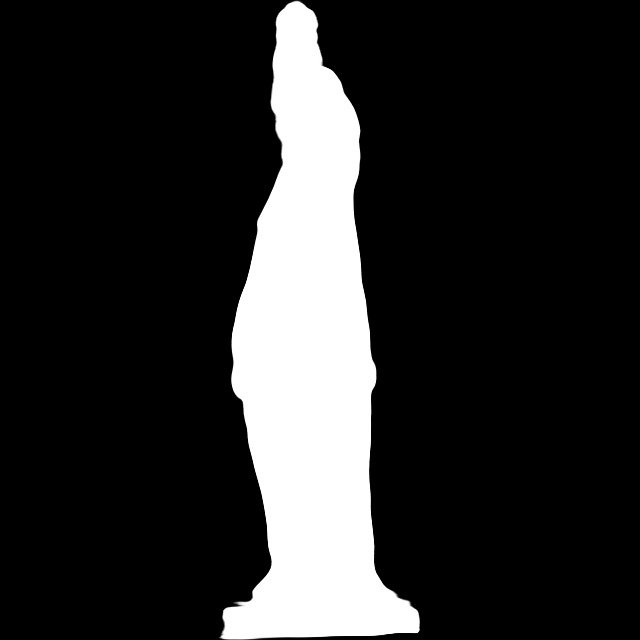}} &
	4 & \raisebox{-.5\height}{\includegraphics[height=14mm,width=14mm,keepaspectratio]{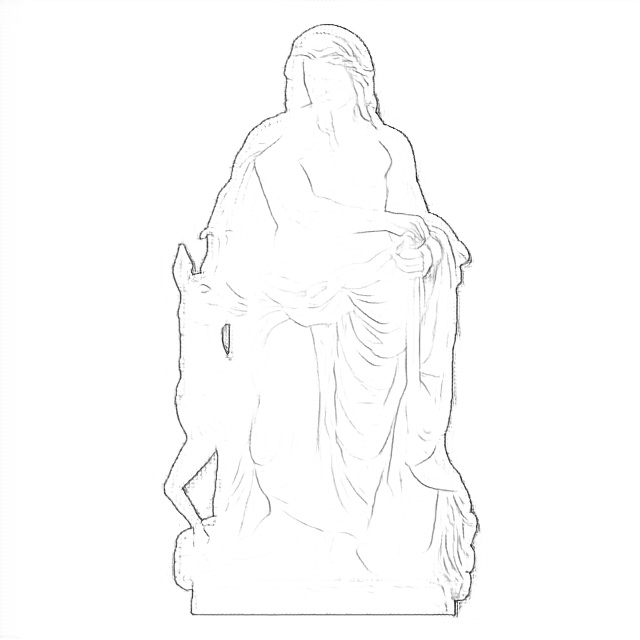}} &  \raisebox{-.5\height}{\includegraphics[height=14mm,width=14mm,keepaspectratio]{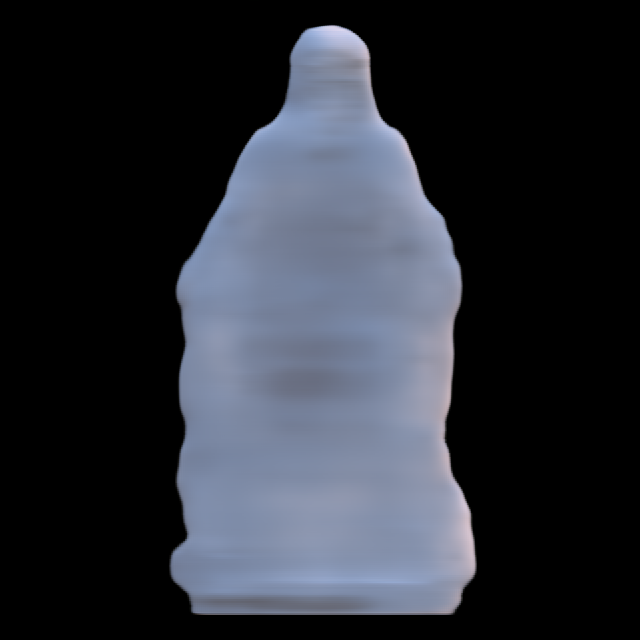}} & \raisebox{-.5\height}{\includegraphics[height=14mm,width=14mm,keepaspectratio]{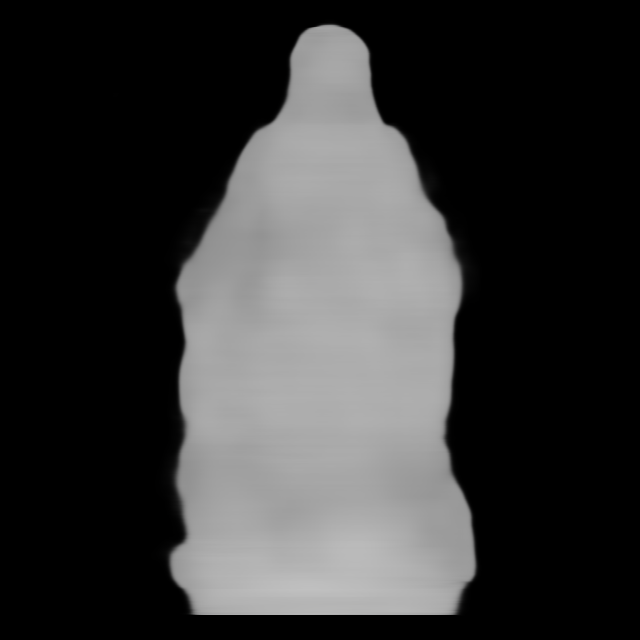}} & \raisebox{-.5\height}{\includegraphics[height=14mm,width=14mm,keepaspectratio]{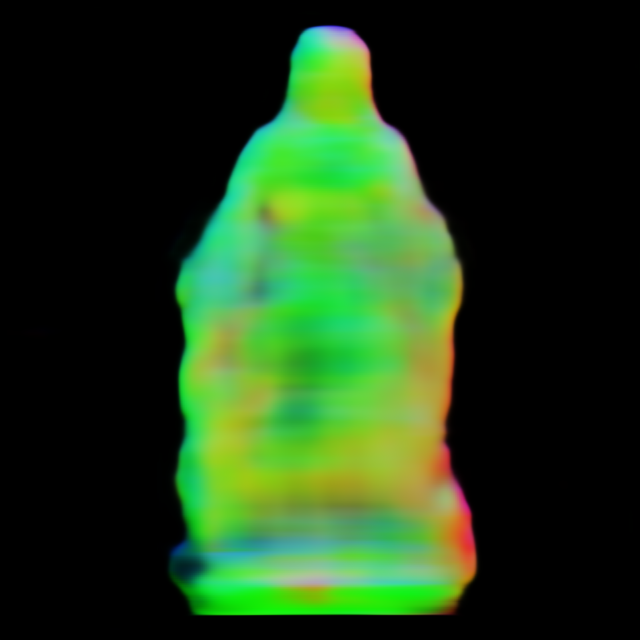}} & \raisebox{-.5\height}{\includegraphics[height=14mm,width=14mm,keepaspectratio]{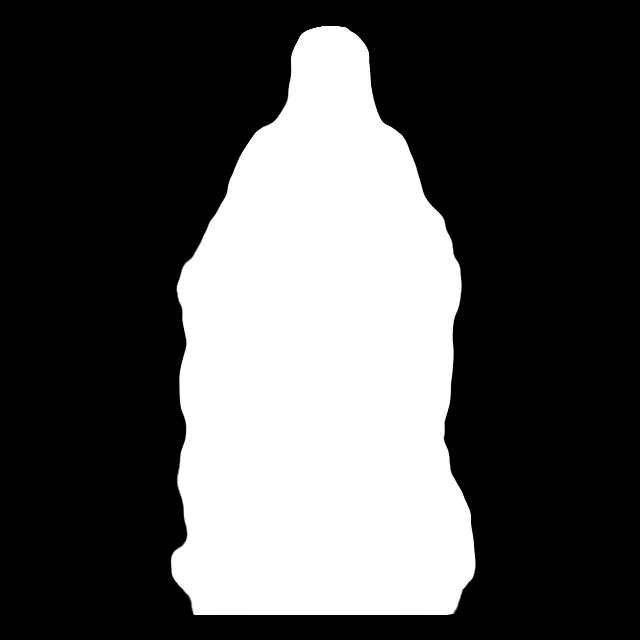}}   \\
	5 & \raisebox{-.5\height}{\includegraphics[height=14mm,width=14mm,keepaspectratio]{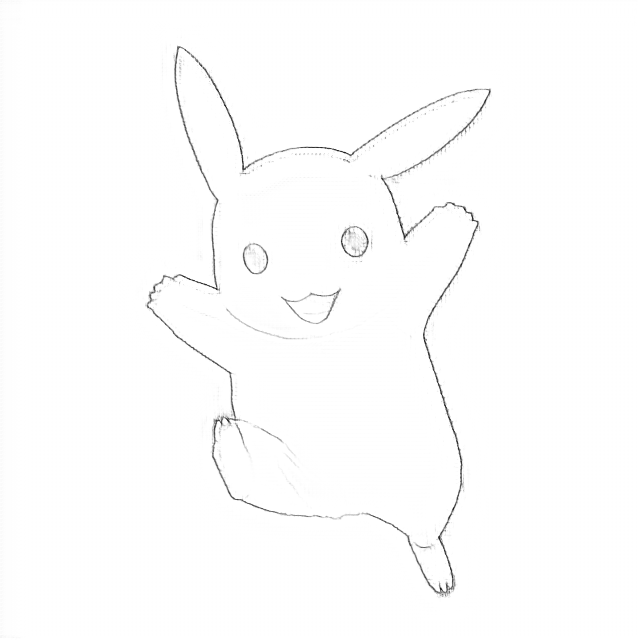}}	&  \raisebox{-.5\height}{\includegraphics[height=14mm,width=14mm,keepaspectratio]{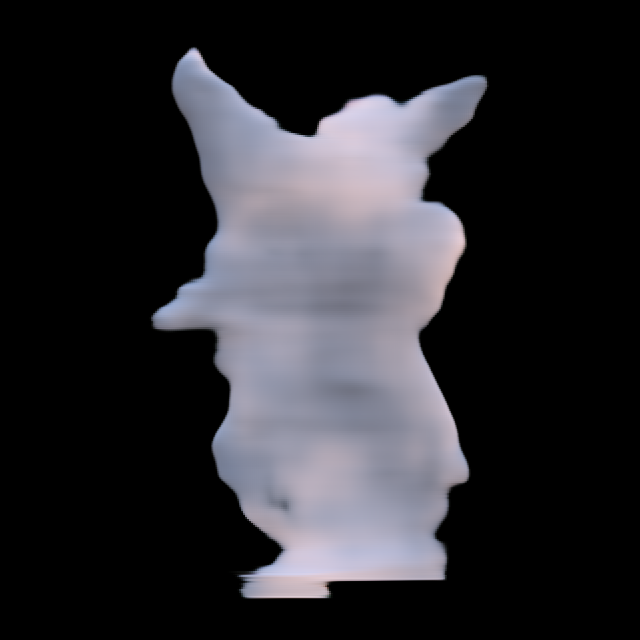}} & \raisebox{-.5\height}{\includegraphics[height=14mm,width=14mm,keepaspectratio]{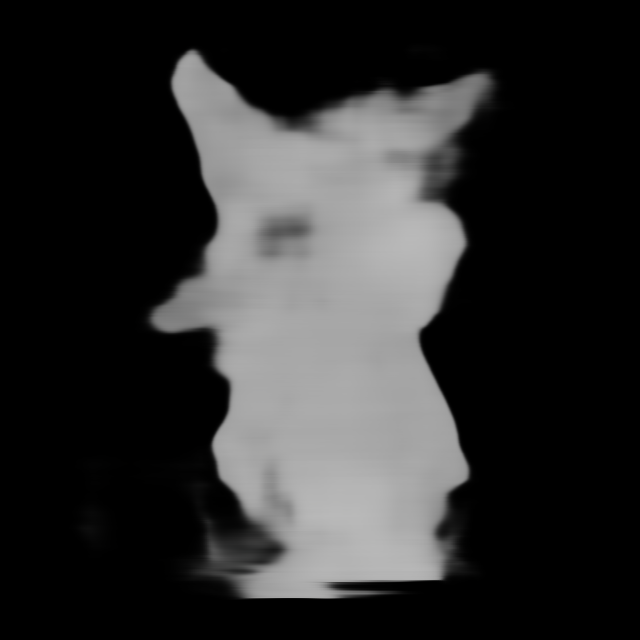}} & \raisebox{-.5\height}{\includegraphics[height=14mm,width=14mm,keepaspectratio]{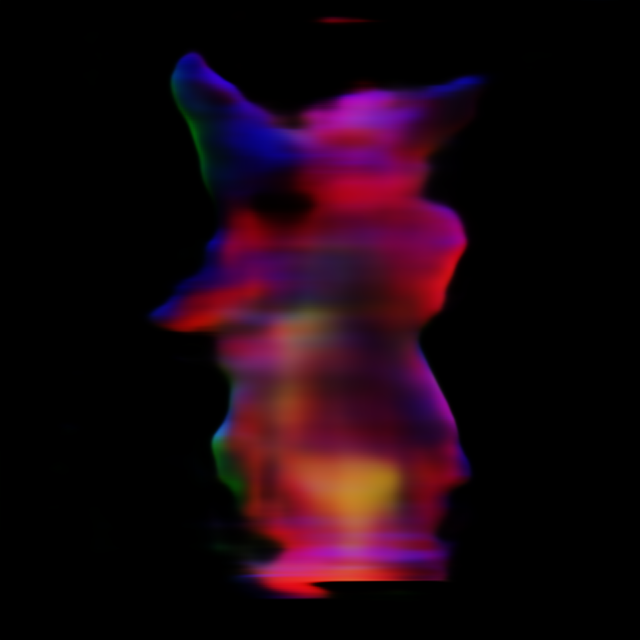}} & \raisebox{-.5\height}{\includegraphics[height=14mm,width=14mm,keepaspectratio]{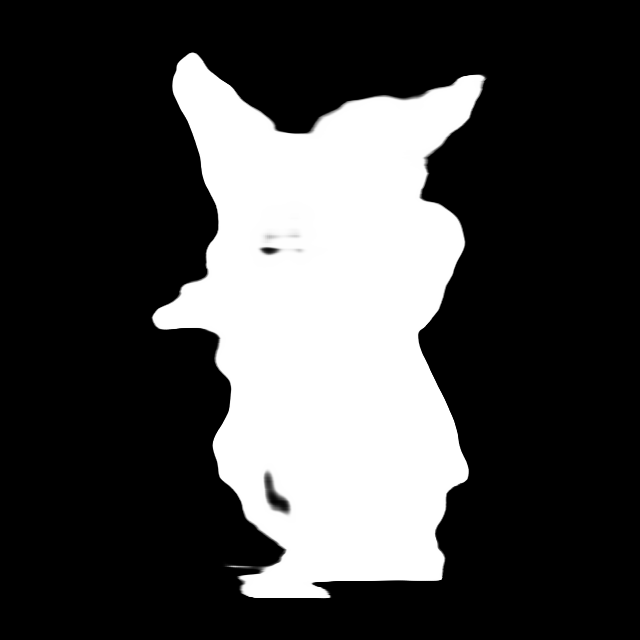}} &
	6 & \raisebox{-.5\height}{\includegraphics[height=14mm,width=14mm,keepaspectratio]{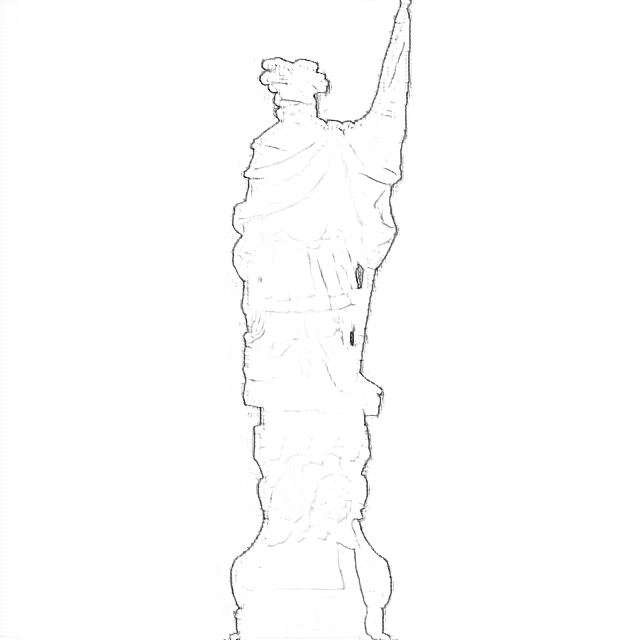}} &  \raisebox{-.5\height}{\includegraphics[height=14mm,width=14mm,keepaspectratio]{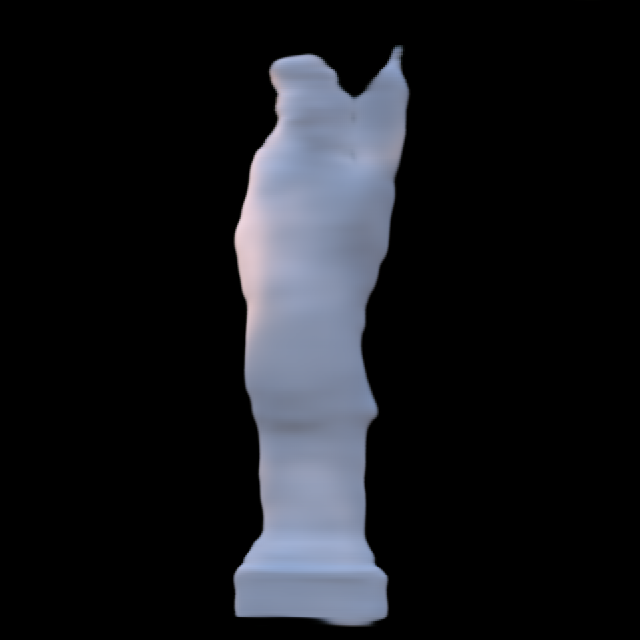}} & \raisebox{-.5\height}{\includegraphics[height=14mm,width=14mm,keepaspectratio]{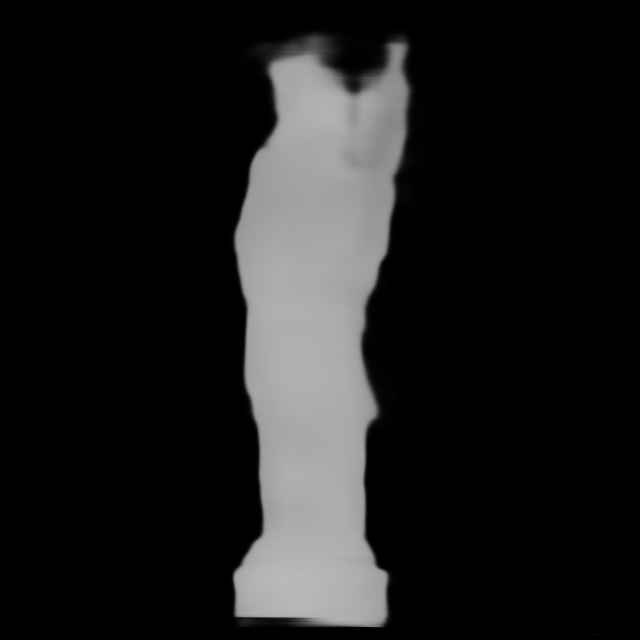}} & \raisebox{-.5\height}{\includegraphics[height=14mm,width=14mm,keepaspectratio]{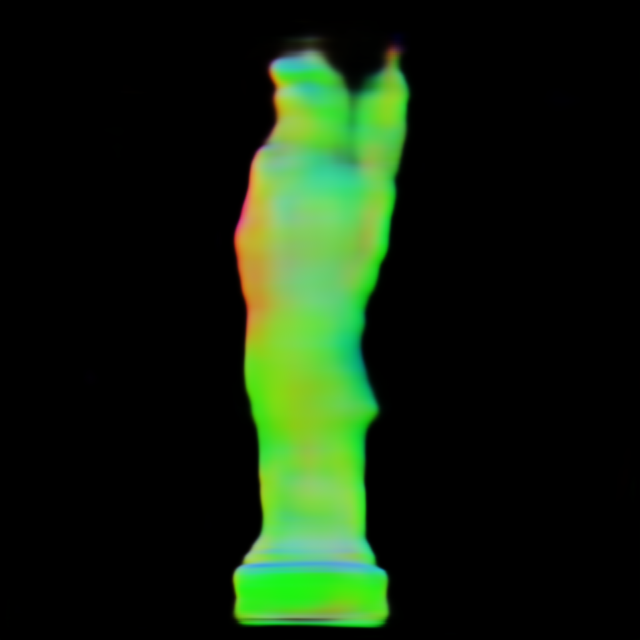}} & \raisebox{-.5\height}{\includegraphics[height=14mm,width=14mm,keepaspectratio]{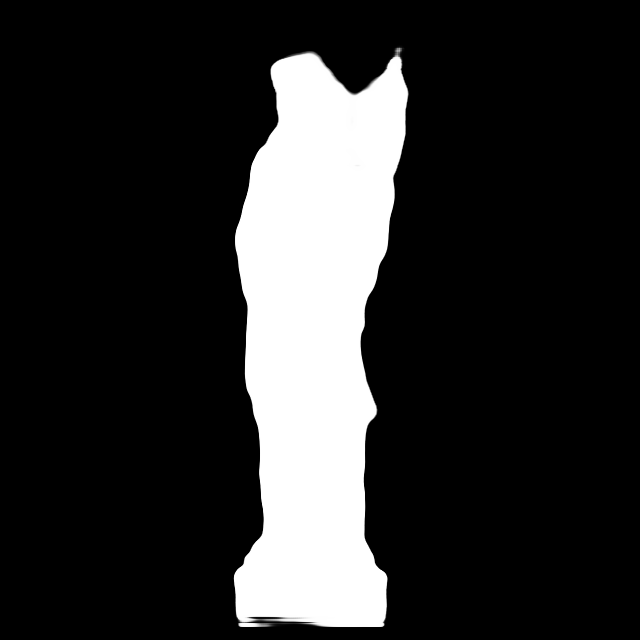}}   \\
	7 & \raisebox{-.5\height}{\includegraphics[height=14mm,width=14mm,keepaspectratio]{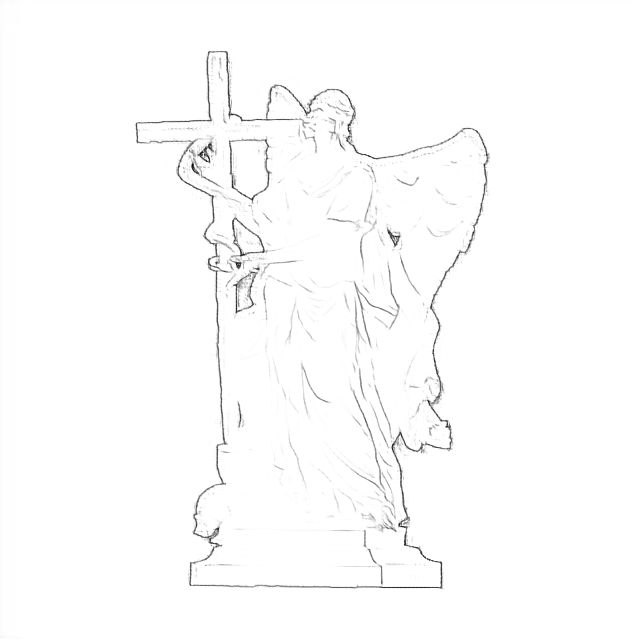}} &  \raisebox{-.5\height}{\includegraphics[height=14mm,width=14mm,keepaspectratio]{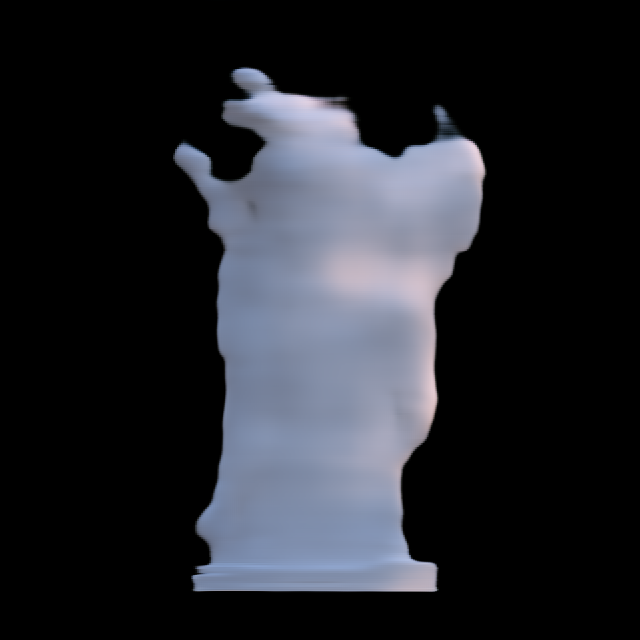}} & \raisebox{-.5\height}{\includegraphics[height=14mm,width=14mm,keepaspectratio]{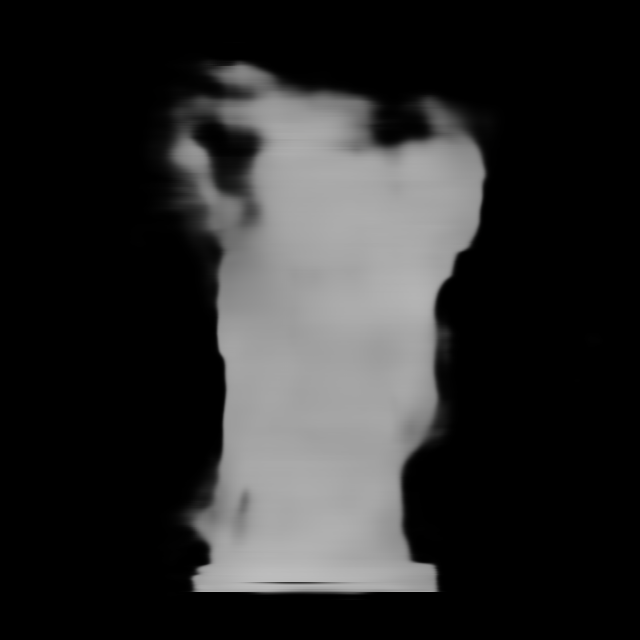}} & \raisebox{-.5\height}{\includegraphics[height=14mm,width=14mm,keepaspectratio]{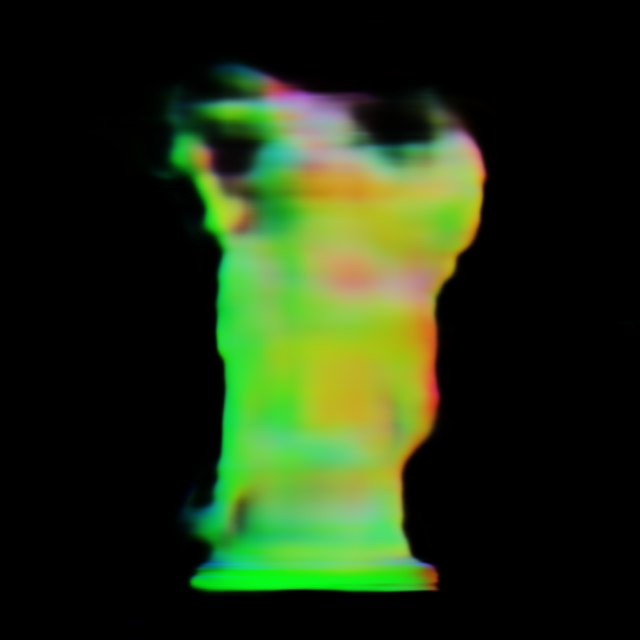}} & \raisebox{-.5\height}{\includegraphics[height=14mm,width=14mm,keepaspectratio]{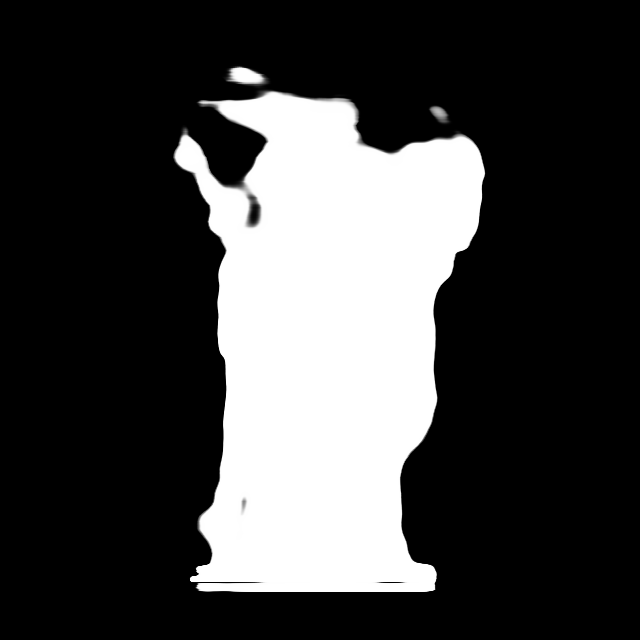}} &
	8 & \raisebox{-.5\height}{\includegraphics[height=14mm,width=14mm,keepaspectratio]{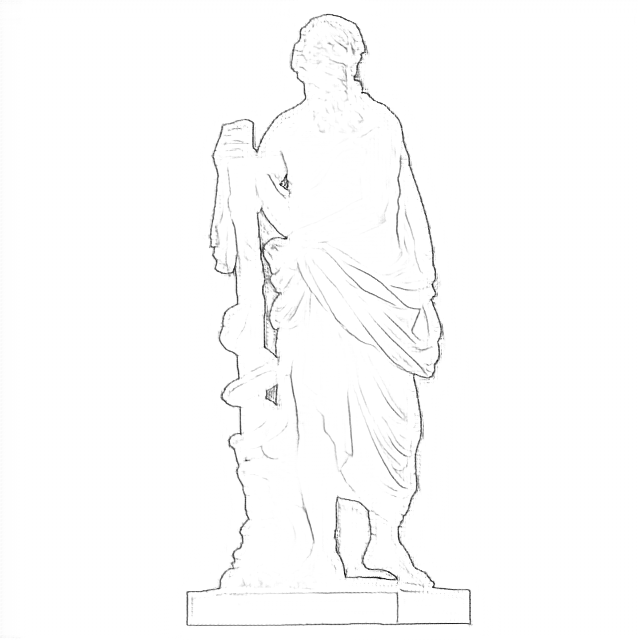}}	&  \raisebox{-.5\height}{\includegraphics[height=14mm,width=14mm,keepaspectratio]{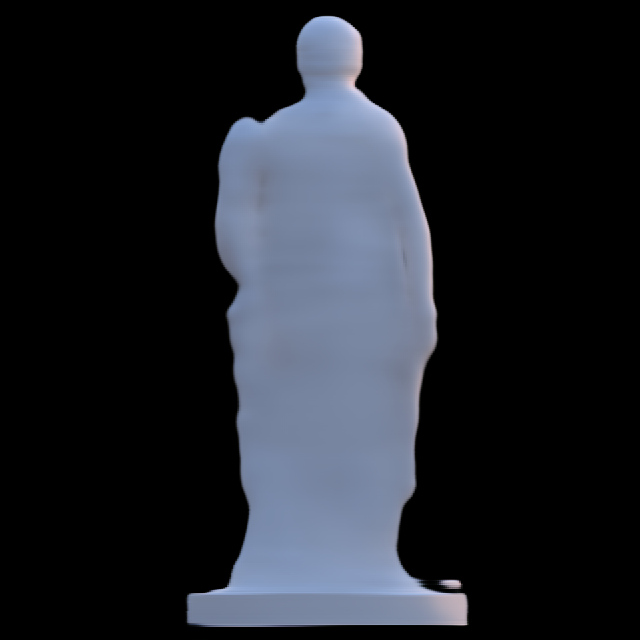}} & \raisebox{-.5\height}{\includegraphics[height=14mm,width=14mm,keepaspectratio]{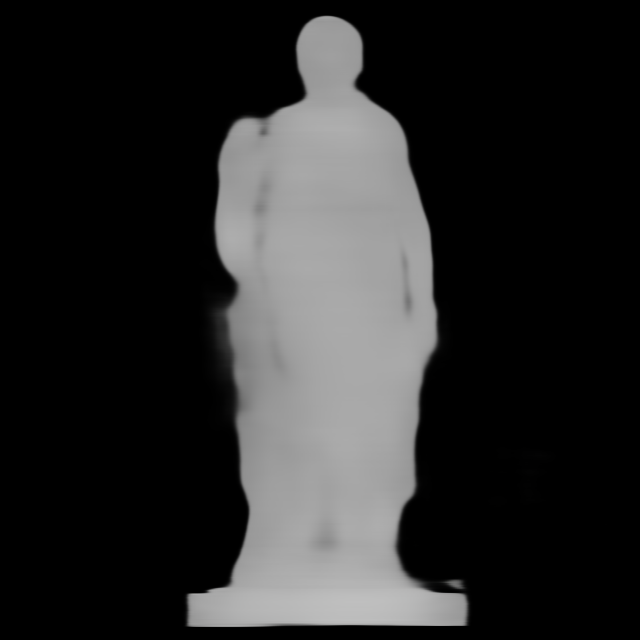}} & \raisebox{-.5\height}{\includegraphics[height=14mm,width=14mm,keepaspectratio]{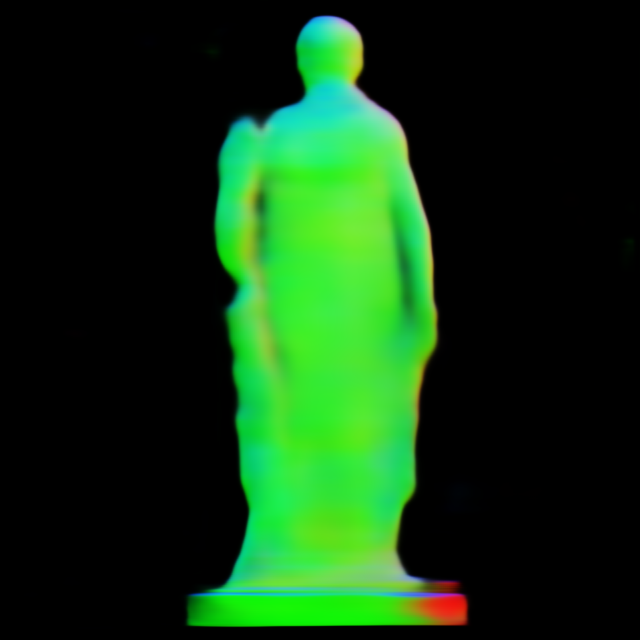}} & \raisebox{-.5\height}{\includegraphics[height=14mm,width=14mm,keepaspectratio]{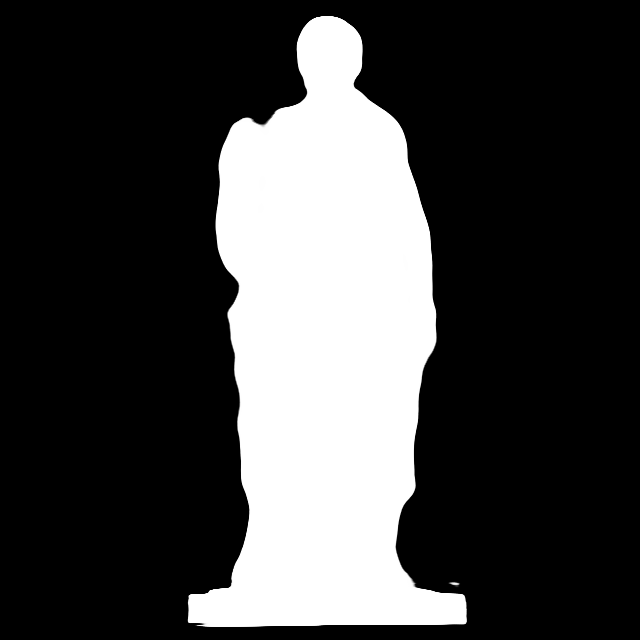}} \\
	\hline
\end{tabular}
\caption{\label{fig:QualitativeResults}
           Preliminary results on our test set of unseen statues. We show the automatically generated sketch, and the outputs of our solution (RGB, depth, normals, mask). (License for Models 1-4, 6-8: \href{https://creativecommons.org/licenses/by-nc/4.0/}{CC BY-NC 4.0}, sketchfab/noe-3d.at; 5: \href{https://creativecommons.org/licenses/by/4.0/}{CC BY 4.0}, sketchfab/HarrisonHag1) }
\end{table*}
\\
We build on the idea of \cite{LearningPCDGen}, which predicts additional views, given an input photograph, estimates a mask plus depth per view and, fusing these predictions, generates a point cloud. Most importantly, by using the projection into 2d for learning, a purely geometric operation, they supervise the learning in an efficient fashion. While we have not yet decided on the final 3d reconstruction method, we adopt the strategy of not directly estimating 3d information. Instead, we require the decoder to come up with intermediary information we intend to use for 3d reconstruction. Also, we extend the capabilities of our network by, additonally to a one-hot image mask separating the figurine from the background, and a depth map, predicting surface normals, as well as an RGB reconstruction. Introducing multiple criterions stabilizes training, while also introducing additional queues to the network without requiring additional inputs at inference time. Altogether this allows us to train completely on synthetic data, not requiring any real data from the target domain. We show outputs of in Table \ref{fig:QualitativeResults}. As for the 3d reconstruction stage, we suggest the use of differentiable rendering techniques of which an evaluation for our use case is in the works. 

\subsection{Data Generation}
We download a set of 110 historical statues from 
\href{www.sketchfab.com}{sketchfab}. We do not filter by epoch, combining ancient greek, ancient roman, and medieval statues. Next, we render 360 orthographic views per statue, rotating 1° per view, using Blender, giving us the RGB, depth, normals, and mask. 

%Since we have only six images of our target sketches, we have to come up with a source of high quality data.
%, circumventing the common problem within cultural heritage, of only having few samples to learn from. 

\subsection{Image to Image Translation}
Only using the gradients of our renderings (canny edge detection, laplacian, or mixture of gaussians) does - in our tests - not suffice to make them indistinguishable to our sinopia images. The remaining domain gap is large enough to make our estimation algorithms work well on our training data, but not on our real world images. Therefore we looked into image-to-image translation techniques. \\
Instead of relying on a fixed function, such as the above mentioned edge filters or similar ones, such as in \cite{pencilFilter}, we turned to style transfer techniques popularized by works such as \cite{styleGatys}\cite{adain}\cite{stylegan}\cite{cyclegan}\cite{pix2pixhd}\cite{style}\cite{cycleganFail}. Using a learned non-linear function greatly increases the expressiveness of our intermediary sketch domain, while preventing a linear 1-to-1 mapping given the right loss function. We rely on the very recent \cite{informativeDrawings} for our sketch generation, which explicitly tackles the encoding of 3d shape via a geometric loss function.

\subsection{Encoder Decoder Network}
To make up for our low data quality we need an expressive model that can be primed with knowledge from many examples, while being able to generalize to unseen statues. Among the possible architectures to address this challenge, (conditional) GANs and autoencoders stand out. We choose to build an encoder-decoder architecture, but add an adversarial component via an additional classification output in the decoder, together with a gradient reversal layer. This technique, as shown in \cite{reversal}\cite{deception}, discourages the encoder to distinguish between different inputs (e.g. different statues), regularizing the weights and increasing generalization between different figurines. Next we describe our architecture in more detail. 
\begin{figure}[b]
  \centering
  \includegraphics[width=0.5\textwidth]{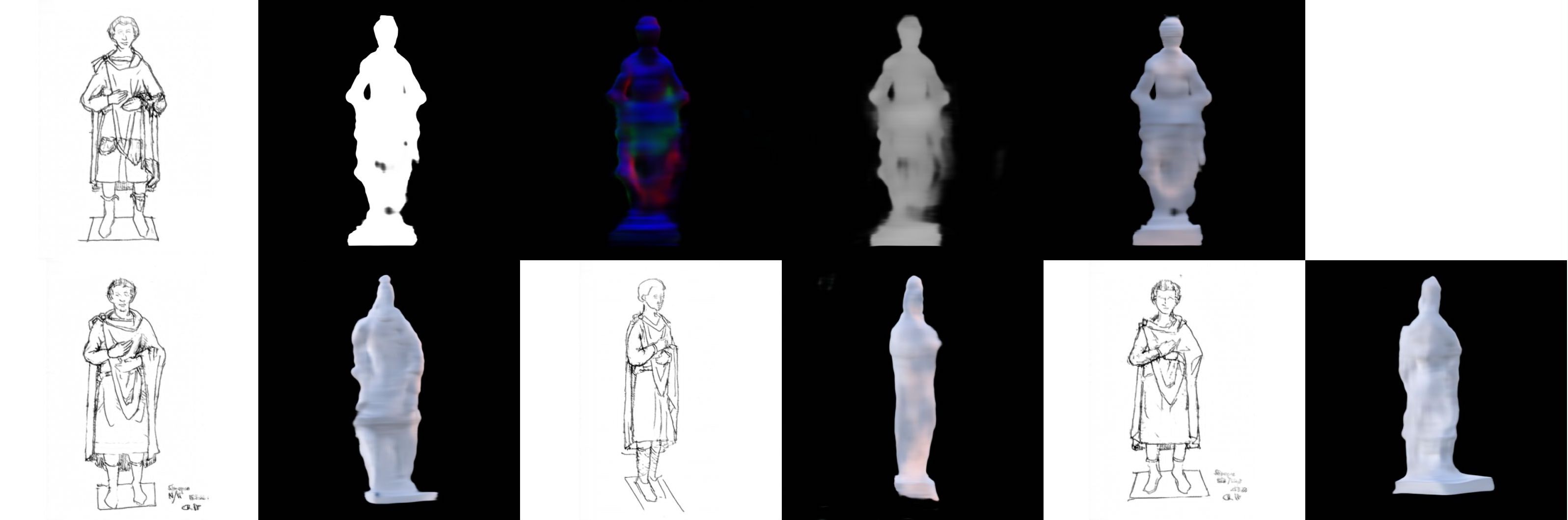}
  \caption{\label{fig:QualitativeResultsSinopia} Preliminary qualitative results on our sinopia. Row 1 shows all outputs of our process, from left to right: First, restoration by an expert (Prof. Dr. Stiegemann), used as input to our network, followed by the mask, normals, depth and color estimate of our approach. Second row shows input-output pairs of additional sketches and their color reconstruction. }
\end{figure}
\textbf{Architectural details}. 
We use an encoder network to extract a powerful representation of a single input sketch depicting the target object. Our encoder leverages a residual design similar to those used in state-of-the-art work \cite{stylegan3} projecting our 640x640 pixel-sized images into a vector space of 2050 dimensions. Similarly, the decoder uses a residual design to generate RGB, depth, normals, and mask images given samples from this latent space, sharing the bigger portion of parameters, before splitting into different heads, one for each output modality and one additional for the classifier, required for our gradient reversal regularization.\\
\textbf{Training details}. 
We use 110 statues to generate 39.600 samples, each consisting of RGB, depth, normals, and mask. We split the set into 91 statues for training (32.760 images), 10 for validation (3.600 images), and 9 for testing (3.240 images). Aside of gradient reversal, we added dropout, and varied our choice as well as scaling of our individual loss terms.

\section{Preliminary Results}
Our network has to generalize to unseen objects in order to solve the problem as presented. Therefore we only test on unseen objects. First, we present preliminary results on our test split, converted renderings, before applying our technique on restorations of historic sinopia. First we present qualitative results (Table \ref{fig:QualitativeResults}) on renderings from unseen objects. Next we evaluate our model qualitatively on six frontal views of our sinopia (Figure \ref{fig:QualitativeResultsSinopia}).
On our synthetic data we see a strong shape reconstruction for statues similar to our training set (humanoid statues from ancient Rome, ancient Greece and the Middle Ages) in all images but 5 and 7. 5, being a modern, non-human figure, strongly differs in shape and style from human statues, 7 contains a lot of unseen detail (the cross). Since details of the training set are transferred to unseen samples, one might want to focus on statues with similar appearance to the target domain. Results on real data show general shape reconstruction, but a lack of detail such as found in the clothing.

\section{Conclusion and Future Work}
We presented a pipeline for 3d geometry reconstruction based on historic drawings. We demonstrated the (semi-)automated reconstruction of RGB, depth, normals, and object mask based only on six medieval sinopia and without requiring any additional sinopia from that or any other time period. We produce all our data via rendering, process the renderings to make them less distinguishable from our target data and finally train an encoder-decoder architecture. Future work encompasses the extension to full 3d reconstruction, based on the predicted information, as well as evaluating the approach on a set of statues consisting only of samples from the era. The second point re-introduces the need for domain knowledge, but should drastically improve the plausability of results. Also, the process as presented can be applied to non-sketch inputs, such as drawings, paintings, (historic) photographs and similar. 
% [] outlook: applicability to non sketch inputs
% - [x] A future development would be a wider training with more types of statues (with morphological features of increasing complexity) to generalise the automatic procedure to a broad variety of historical statues. The authors could mention also this direction in the future work.

\bibliographystyle{eg-alpha-doi}  
\bibliography{egbibsample}     

\end{document}